# Optimization of measurement configurations for geometrical calibration of industrial robot

Alexandr Klimchik[1,2], Anatol Pashkevich[1,2], Yier Wu[1,2],
Benoît Furet[2], Stephane Caro[2]

[1] Ecole des Mines de Nantes, 4 rue Alfred-Kastler, Nantes 44307, France
[2] Institut de Recherche en Communications et Cybernétique de Nantes, 44321 France
{alexandr.klimchik, anatol.pashkevich, yier.wu}@mines-nantes.fr,
{benoit.furet, stephane.caro}@irccyn.ec-nantes.fr

**Abstract.** The paper is devoted to the geometrical calibration of industrial robots employed in precise manufacturing. To identify geometric parameters, an advanced calibration technique is proposed that is based on the non-linear experiment design theory, which is adopted for this particular application. In contrast to previous works, the calibration experiment quality is evaluated using a concept of the user-defined test-pose. In the frame of this concept, the related optimization problem is formulated and numerical routines are developed, which allow user to generate optimal set of manipulator configurations for a given number of calibration experiments. The efficiency of the developed technique is illustrated by several examples.

## 1 Introduction

In the usual engineering practice, the accuracy of robotic manipulator depends on a number of factors. Usually, for the industrial applications where the external forces/torques applied to the end-effector are relatively small, the prime source of the manipulator inaccuracy is the *geometrical errors*, which are responsible for about 90% of the total position error [1]. These errors are associated with the differences between the nominal and actual values of the link/joint parameters. Typical examples of them are the differences between the nominal and the actual length of links, the differences between zero values of actuator coordinates in the real robot and the mathematical model embedded in the controller (joint offsets) [2]. They can be also induced by the non-perfect assembling of different links and lead to shifting and/or rotation of the frames associated with different elements, which are normally assumed to be matched and aligned. It is clear that the errors in geometrical parameters do not depend on the manipulator configuration, while their effect on the position accuracy depends on the last one. At present, there exists various sophisticated calibration techniques that are able to identify the differences between the actual and the nominal geometrical parameters [3,4], however the problem of optimal selection of measurement configurations is still in the focus of robotic experts.

The primary motivation of this research area is the possibility of essential reduction the measurement error impact. From point of view of classical experiment design theory [5] this goal can be achieved by proper selection of measurement poses that differ from each other as much as possible. However, in spite of potential advantages of this approach and potential benefits to improve the identification accuracy significantly, only few works addressed to the issue of the best measurement pose selection [6-8]. Related works focus on optimization of some abstract performance measures [9-12] (condition number of the aggregated Jacobian matrix, its determinant, etc.) which are not directly related to the robot precision for particular industrial application. In contrast, this work operates with an industry-oriented performance measure that is directly related to the robot position accuracy in a given workspace location (corresponding to so-called test configuration). Using this idea, in the following sections the problem of calibration experiment design is formulated as a constrained optimization problem (taking into account some specific technological requirements) and is solved for serial manipulators with 2 and 6 degrees of freedoms.

## 2 Problem of geometrical calibration

Let us consider a serial robotic manipulator whose end-effector position $\mathbf{p}$ is computing using the geometrical model

$$\mathbf{p} = g(\mathbf{q}, \mathbf{\Pi}) \tag{1}$$

which includes the vector of the unknown parameters $\mathbf{\Pi}$ to be identified and where the vector $\mathbf{q}$ aggregates all joint coordinates. Usually the most essential components of the vector $\mathbf{\Pi}$ are the deviations of the robot link lengths $l_i$ and the offsets $\Delta q_j$ in the actuated joints, but in some cases it may also include inclinations of the joint axes, etc. In practice, the above defined function $g(.)$ can be extracted from the product of homogeneous transformation matrices

$$\mathbf{T} = \mathbf{T}_{base} \left( \prod_{i=1}^{n} \mathbf{T}_i(q_i, \mathbf{\Pi}_i) \right) \mathbf{T}_{tool} \tag{2}$$

which are widely used in robotic kinematics. Here, $\mathbf{T}_{base}$ and $\mathbf{T}_{tool}$ denote the 'Base' and 'Tool' transformations respectively, $\mathbf{T}_i(q_i, \mathbf{\Pi}_i)$ defines transformations related to the $i$-th actuated joint. Here, $\mathbf{T}$, $\mathbf{T}_i$, $\mathbf{T}_{base}$, $\mathbf{T}_{tool}$ are $4 \times 4$ matrices that are computed as a product of simple translation/rotation matrices, for which the number of multipliers and their order is defined by robot geometrical model. Since the deviations of geometrical parameters $\Delta \mathbf{\Pi}$ are usually relatively small, calibration usually relies on the linearized model [8]

$$\mathbf{p} = g(\mathbf{q}, \mathbf{\Pi}_0) + \mathbf{J}(\mathbf{q}, \mathbf{\Pi}_0) \Delta \mathbf{\Pi} \tag{3}$$

which includes the Jacobian $\mathbf{J}(\mathbf{q}, \mathbf{\Pi}_0) = \partial g(\mathbf{q}, \mathbf{\Pi}_0)/\partial \mathbf{\Pi}$ computed for the nominal parameters $\mathbf{\Pi}_0$.

In the frame of this work, the following assumptions concerning the manipulator model and the measurement equipment limitations are accepted:

**A1**: *each calibration experiment produces two vectors* $\{\mathbf{p}_i, \mathbf{q}_i\}$, which define the robot end-effector position and corresponding joint angles;

**A2**: the calibration relies on the *measurements of the end-effector position only* (i.e. Cartesian coordinates x, y, z; such approach allows us to avoid the problem of different units and to use three points with position instead of one with position and orientation)*;*

**A3:** *the measurements errors* are treated as independent identically distributed random values with zero expectation and standard deviation $\sigma$.

Because of the measurement errors, the unknown parameters $\mathbf{\Pi}$ are always identified approximately and their estimates $\hat{\mathbf{\Pi}}$ can be also treated as random values. For this reason, the "identification quality" is usually evaluated via the covariance matrix $\mathrm{cov}(\hat{\mathbf{\Pi}})$, whose elements should be as small as possible. However, this approach does not provide the final user with a clear engineering characteristic of the accuracy improvement, which is achieved due to calibration. Thus, it is proposed to use another performance measure that directly evaluates the robot accuracy after compensation of the geometrical errors, which in the frame of the adopted above notations can be expressed as.

$$\boldsymbol{\varepsilon}_p(\mathbf{q}) = g(\mathbf{q}, \hat{\mathbf{\Pi}}) - g(\mathbf{q}, \mathbf{\Pi}) \tag{4}$$

where $\mathbf{q}$ defines the manipulator configuration. Further, to take into account particularities of the considered technological application, it is reasonable to limit the possible configurations set by a single one $\mathbf{q}_0$, which is treated as a typical for the manufacturing task. It is obvious that definition of $\mathbf{q}_0$ ("test-pose") is a non-trivial step that completely relies on the user experience and his/her understanding of the technological process. The main substantiation for this approach is to take into account that all geometrical errors have different influence on the end-effector position and this influence varies throughout the workspace. However, in practice, high accuracy is required in the neighborhood of the prescribe trajectory only.

Taking into account that the geometric parameter estimate $\hat{\mathbf{\Pi}}$ is computed via the best fitting of the data set $\{\mathbf{p}_i, \mathbf{q}_i\}$ by the function (1), the expectation of the position errors after compensation is equal to zero, i.e. $E(\boldsymbol{\varepsilon}_p) = 0$. However, the standard deviation $E(\boldsymbol{\varepsilon}_p^T \boldsymbol{\varepsilon}_p)$ essentially depends on the measurement configurations (which, from point of view of the experiment design theory, can be treated as the plan of the calibration experiments). This allows us to present the considered problem in the following way:

**Problem:** *For a given number of experiments* $m$, *find a set of measurement configurations* $\{\mathbf{q}_1, \ldots \mathbf{q}_m\}$ *defined by the vectors of the joint variables* $\mathbf{q}_i$ *that ensures minimum value of the position error s.t.d. for the test configuration* $\mathbf{q}_0$:

$$E\left(\| g(\mathbf{q}_0, \hat{\mathbf{\Pi}}) - g(\mathbf{q}_0, \mathbf{\Pi}) \|^2 \right) \to \min_{\{\mathbf{q}_1 \ldots \mathbf{q}_m\}} \tag{5}$$

*where* $\|\varepsilon_p\|$ *denotes the Euclidian norm of the vector* $\varepsilon_p$.

In the following sections this optimization problem will be solved subject to the additional constraints imposed by the application area.

## 3  Influence of measurement errors

For comparison purpose, let us first evaluate the influence of the measurement errors on the accuracy of the geometrical parameters identification. Using the linear approximation of the geometrical model (3), the deviation of the desired parameters with respect to their nominal values $\Delta \Pi = \hat{\Pi} - \Pi_0$ can be obtained from the minimum least-square formulation

$$\sum_{i=1}^{m} \left( \mathbf{J}_i \Delta \Pi - \Delta \mathbf{p}_i \right)^T \left( \mathbf{J}_i \Delta \Pi - \Delta \mathbf{p}_i \right) \to \min_{\Delta \Pi} \qquad (6)$$

which yields expression

$$\Delta \hat{\Pi} = \left( \sum_{i=1}^{m} \mathbf{J}_i^T \mathbf{J}_i \right)^{-1} \cdot \left( \sum_{i=1}^{m} \mathbf{J}_i^T \Delta \mathbf{p}_i \right) \qquad (7)$$

where $\Delta \mathbf{p}_i = \mathbf{p}_i - \mathbf{p}_{0i}$ denotes the shift of the end-effector position $\mathbf{p}_i$ for the *i*-th experiment with respect to the location corresponding to the nominal geometrical parameters $\Pi_0$ and measurement configuration $\mathbf{q}_i$. To increase the identification accuracy, the foregoing linearized procedure has to be applied several times, in accordance with the following iterative algorithm:

**Step 1.** Carry out experiments and collect the input data in the vectors of generalized coordinates $\mathbf{q}_i$ and end-effector position $\mathbf{p}_i$. Initialize $\Delta \Pi = 0$.

**Step 2.** Compute end-effector position via direct kinematic model (1) using initial generalized coordinates $\mathbf{q}_i$

**Step 3.** Compute residuals and unknown parameters $\Delta \Pi$ via (7)

**Step 4.** Modify mathematical model and generalized coordinates $\Pi$ and $\mathbf{q}_i$.

**Step 5.** If required accuracy is not satisfied, repeat from Step 2.

Further, to integrate the measurement errors $\varepsilon_i$ in equation (7), $\Delta \mathbf{p}_i$ can be expressed as

$$\Delta \mathbf{p}_i = \mathbf{J}_i \Delta \Pi + \varepsilon_i \qquad (8)$$

where $\varepsilon_i$ are assumed to be independent identically distributed (i.i.d.) random values with zero expectation $\mathrm{E}(\varepsilon_i) = \mathbf{0}$ and the variance $\mathrm{E}(\varepsilon_i^T \varepsilon_i) = \sigma^2$. Hence, as follows from (7), the geometric parameters estimate $\hat{\Pi}$ can be presented as the sum

$$\hat{\Pi} = \Pi_0 + \left( \sum_{i=1}^{m} \mathbf{J}_i^T \mathbf{J}_i \right)^{-1} \left( \sum_{i=1}^{m} \mathbf{J}_i^T \varepsilon_i \right) \qquad (9)$$

where the first term corresponds to the expectation of this random variable. From the latter expression, the covariance matrix of $\hat{\mathbf{\Pi}}$, which defines the identification accuracy, can be computed as

$$\text{cov}(\hat{\mathbf{\Pi}}) = \left(\sum_{i=1}^{m} \mathbf{J}_i^T \mathbf{J}_i\right)^{-1} \text{E}\left(\sum_{i=1}^{m} \mathbf{J}_i^T \boldsymbol{\varepsilon}_i \boldsymbol{\varepsilon}_i^T \mathbf{J}_i\right) \left(\sum_{i=1}^{m} \mathbf{J}_i^T \mathbf{J}_i\right)^{-1}. \tag{10}$$

So, considering that $\text{E}\left(\boldsymbol{\varepsilon}_i \boldsymbol{\varepsilon}_i^T\right) = \sigma^2 \mathbf{I}$, the desired covariance matrix can be simplified to:

$$\text{cov}(\hat{\mathbf{\Pi}}) = \sigma^2 \left(\sum_{i=1}^{m} \mathbf{J}_i^T \mathbf{J}_i\right)^{-1} \tag{11}$$

where $\sigma$ is the s.t.d. of the measurement errors. Hence, the impact of the measurement errors on the identified values of the geometric parameters is defined by the matrix sum $\sum_{i=1}^{m} \mathbf{J}_i^T \mathbf{J}_i$ that is also called the information matrix.

It should be stressed that most of the related works [9-11] reduce the calibration experiment design problem to the problem of covariance matrix minimization, which is evaluated by means of the determinant, Euclidian norm, trace, singular values, etc. However, because of some essential disadvantages mentioned in the previous section, this approach may provide a solution, which does not guarantee the best position accuracy for typical manipulator configurations defined by the manufacturing process. This motivates another approach presented below.

## 4  Test-pose based approach

To overcome the above mentioned difficulty, it is prudent to introduce another performance measure, which is directly related to the robot accuracy after compensation of the geometrical errors. Besides, to take into account that the desired accuracy should be achieved for rather limited workspace area, it is proposed to limit possible manipulator configurations by a single one (corresponding to joint variables $\mathbf{q}_0$), which further will be referred to as a test-pose. It is evident that this performance measure is attractive for practicing engineers and also allows to avoid the multiobjective optimization problem that arises while minimizing all elements of the covariance matrix (11) simultaneously. In addition, using this approach, it is possible to find a balance between accuracy of different geometrical parameters whose influence on the final robot accuracy is unequal.

In more formal way, the proposed performance measure $\rho_0$ may be defined as the s.t.d. of the distance between the desired end-effector position and its real position achieved after application of the geometrical error compensation technique.

Using the notations from the previous section, this distance may be computed as the Euclidean norm of the vector $\delta\mathbf{p} = \mathbf{J}_0 \delta\mathbf{\Pi}$, where the subscript '0' is related to the test pose $\mathbf{q}_0$ and $\delta\mathbf{\Pi} = \hat{\mathbf{\Pi}} - \mathbf{\Pi}$ is the difference between the estimated and true values of the robot geometrical parameters. It can be proved that the above presented identification algorithm provides an unbiased compensation, i.e. $\mathrm{E}(\delta\mathbf{p}) = \mathbf{0}$, while the standard deviation of the compensation error $\mathrm{E}(\delta\mathbf{p}^T \delta\mathbf{p})$ can be expressed as

$$\rho_0^2 = \mathrm{E}\left(\delta\mathbf{\Pi}^T \mathbf{J}_0^T \mathbf{J}_0 \delta\mathbf{\Pi}\right) \qquad (12)$$

Taking into account geometrical meaning of $\rho_0$, this value will be used as a numerical measure of the error compensation quality (and also as a quality measure of the related plan of calibration experiments). This expression can be simplified by presenting the term $\delta\mathbf{p}^T \delta\mathbf{p}$ as the trace of the matrix $\delta\mathbf{p}\delta\mathbf{p}^T$, which yields

$$\rho_0^2 = \mathrm{trace}\left(\mathbf{J}_0 \, \mathrm{E}\left(\delta\mathbf{\Pi}\delta\mathbf{\Pi}^T\right) \mathbf{J}_0^T\right) \qquad (13)$$

Further, taking into account that $\mathrm{E}(\delta\mathbf{\Pi}\delta\mathbf{\Pi}^T)$ is the covariance matrix of the geometrical parameters estimate $\hat{\mathbf{\Pi}}$, the proposed performance measure (13) can be presented in the final form as

$$\rho_0^2 = \sigma^2 \, \mathrm{trace}\left(\mathbf{J}_0 \left(\sum_{i=1}^{m} \mathbf{J}_i^T \mathbf{J}_i\right)^{-1} \mathbf{J}_0^T\right) \qquad (14)$$

As follows from this expression, the proposed performance measure $\rho_0^2$ can be treated as the weighted trace of the covariance matrix (11), where the weighting coefficients are computed using the test pose coordinates. It has obvious advantages compared to previous approaches, which operate with "pure" trace of the covariance matrix and involve straightforward summing of the covariance matrix diagonal elements, which may be of different units.

Using this performance measure, the problem of calibration experiment design can be reduced to the following optimization problem

$$\mathrm{trace}\left(\mathbf{J}_0 \left(\sum_{i=1}^{m} \mathbf{J}_i^T \mathbf{J}_i\right)^{-1} \mathbf{J}_0^T\right) \to \min_{\{\mathbf{q}_1 \ldots \mathbf{q}_m\}} \qquad (15)$$

whose solution gives a set of the desired manipulator configurations $\{\mathbf{q}_1, \ldots \mathbf{q}_m\}$. It is evident that here an analytical solution can hardly be obtained, so a numerical approach is the only reasonable one. An application of this approach for the design of the manipulator calibration experiments and its advantages are illustrated below. Geometrical interpretation of the proposed approach is presented in Fig. 1, where the performance measure $\rho_0$ defines the position error for the target point after calibration.

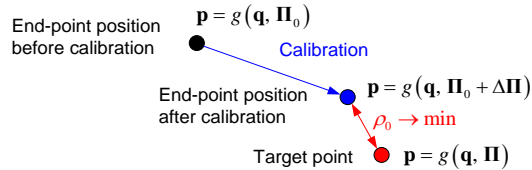

**Fig. 1.** Geometrical interpretation of the test-pose based approach

## 5 Illustrative example

Let us illustrate the advantages of the test-pose-based approach by an example of the geometrical calibration of the 2-link manipulator. For this manipulator, the end-effector position can be computed as

$$x = (l_1 + \Delta l_1)\cos q_1 + (l_2 + \Delta l_2)\cos(q_1 + q_2)$$
$$y = (l_1 + \Delta l_1)\sin q_1 + (l_2 + \Delta l_2)\sin(q_1 + q_2)$$
(16)

where $x$ and $y$ define the end-effector position, $l_1, l_2$ and $\Delta l_1, \Delta l_2$ are nominal link lengths and their deviations (that should be identified), $q_1$, $q_2$ are the joint coordinates that define manipulator configuration. It can be proved that, in the case of $\Delta \Pi = (\Delta l_1, \Delta l_2)$ the parameter covariance matrix does not depend on the angles $q_{1i}$ and can be expressed as:

$$\text{cov}(\Delta \hat{\Pi}) = \frac{\sigma^2}{m^2 - \left(\sum_{i=1}^{m}\cos q_{2i}\right)^2} \begin{bmatrix} m & -\sum_{i=1}^{m}\cos q_{2i} \\ -\sum_{i=1}^{m}\cos q_{2i} & m \end{bmatrix}$$
(17)

where $m$ is the number of experiments and $i = 1,...m$.

For comparison purposes the design of experiment problem was solved using both the known approaches and the proposed one. It can be shown that here it is not reasonable to use the A-criterion (the goal of A-criterion is to minimize the trace of the covariance matrix) because the trace of the relevant information matrix does not depend on the plan of experiments. Further, it was proved that the criteria that operate with the covariance matrix determinant (D and D* criteria, [8]; the goal of D-criterion is to minimize the determinant of the covariance matrix, the goal of D*-criterion is to ensure independence of the identified parameters and to minimize the determinant of the covariance matrix that is diagonal) lead to minimization of $\left|\sum_{i=1}^{m}\cos q_{2i}\right|$. This solution provides good accuracy on average, but not for the test configuration $(q_{10}, q_{20})$.

For the proposed performance measure $\rho_0^2$, the basic expression (14) can be transformed to

$$\rho_0^2 = 2\sigma^2 \left( m - \cos q_{20}\sum_{i=1}^{m}\cos q_{2i} \right) \bigg/ \left( m^2 - \left(\sum_{i=1}^{m}\cos q_{2i}\right)^2 \right)$$
(18)

Here, the minimum value of $\rho_0^2$ is achieved when

$$\sum_{i=1}^{m}\cos q_{2i} = m\left(1 - |\sin q_{20}|\right)\big/\cos q_{20}$$
(19)

and is equal to

$$\rho_{0\min}^2 = \left(\sigma^2/m\right)\cos^2 q_{20}\Big/\left(1-\left|\sin q_{20}\right|\right) \qquad (20)$$

It is evident that general solution of equation (19) for $m$ configurations can be replaced by the decomposition of the whole configuration set by the subsets of 2 and 3 configurations (while providing the same identification accuracy). This essentially reduces computational complexity and allows user to reduce number of different configurations without loss of accuracy.

Compared with other approaches, it should be mentioned that in the test pose $(q_{10}, q_{20})$, the D-criterion insures the accuracy $\rho_D^2 = 2\sigma^2/m$ only. Corresponding loss of the accuracy is presented in Table 1. It is shown that the test-pose based approach allows us to improve the accuracy of the end-effector position up to 41%.

**Table 1.** Accuracy comparison for D-based and test-pose based approaches.

| $\left|q_{20}\right|$, deg | 0° | 30° | 60° | 90° | 120° | 150° | 180° |
|---|---|---|---|---|---|---|---|
| $\rho_D^2/\sigma^2$ | 1 | 1 | 1 | 1 | 1 | 1 | 1 |
| $\rho_0^2/\sigma^2$ | 0.5 | 0.75 | 0.83 | 1 | 0.83 | 0.75 | 0.5 |
| $\rho_D/\rho_0$, % | 41 | 15 | 10 | 0 | 10 | 15 | 41 |

To illustrate advantages of the proposed approach, Fig. 2 presents three plots showing geometrical error compensation efficiency for different calibration plans. These results correspond to the manipulator parameters $l_1 = 1\,\text{m}$, $l_2 = 0.8\,\text{m}$, two measurement configurations $m = 2$, the test pose $\mathbf{q}_0 = (-45°, 20°)$, and s.t.d. of the measurement errors $\sigma = 10^{-3}\,\text{m}$. The calibration experiment has been repeated 100 times. In the case (a), the plan of experiments corresponds to $\mathbf{q}_1 = (0°, -10°)$ and $\mathbf{q}_2 = (0°, 10°)$. In the case (b), the measurement configurations are $\mathbf{q}_1 = (0°, -90°)$ and $\mathbf{q}_2 = (0°, 90°)$ and insure that $\sum_{i=1}^{2}\cos q_{2i} = 0$. And for the case (c), the measurement configurations $\mathbf{q}_1 = (0°, -46°)$ and $\mathbf{q}_2 = (0°, 46°)$ were computed using equation (19). These results show that the proposed approach allows us to increase accuracy of the end-point location on average by 18% comparing to the calibration using D-optimal plan and by 48% comparing to the calibration using non-optimal plan.

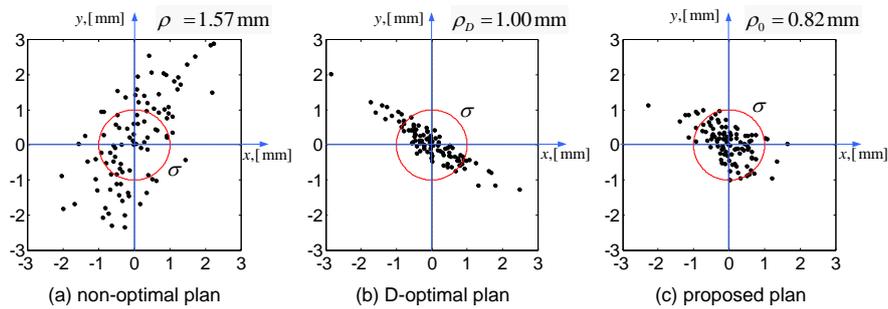

(a) non-optimal plan  (b) D-optimal plan  (c) proposed plan

**Fig. 2.** The accuracy of geometrical error compensation for different plans of calibration experiments: identification of parameters $\Delta l_1, \Delta l_2$ for measurement errors with $\sigma = 10^{-3}\,\text{m}$

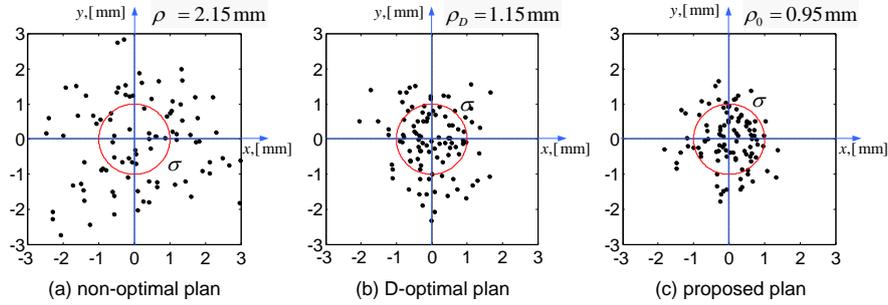

(a) non-optimal plan  (b) D-optimal plan  (c) proposed plan

**Fig. 3.** The accuracy of geometrical error compensation for different plans of calibration experiments: identification of parameters $\Delta l_1, \Delta l_2, \Delta q_1, \Delta q_2$ for measurement errors with $\sigma = 10^{-3}$ m

In the frame of this example, it was also studies the case of the joint offsets calibration, where $\Delta \Pi = (\Delta q_1, \Delta q_2)$. It has been proved that expressions (18)-(20) are also valid in this case. This allows us to suggest a hypothesis that a more general case of simultaneous calibration of the link lengths and joint offsets $\Delta \Pi = (\Delta l_1, \Delta l_2, \Delta q_1, \Delta q_2)$ can be also solved using the same expressions. This hypothesis has been confirmed by the simulation results presented in Fig. 3, where calibration was based on three measurements ($m = 3$). Here, case (a) employees the configurations $\mathbf{q}_1 = (0°, -10°)$, $\mathbf{q}_2 = (0°, 0°)$ and $\mathbf{q}_3 = (0°, 10°)$; case (b) uses the configurations $\mathbf{q}_1 = (0°, -120°)$, $\mathbf{q}_2 = (0°, 0°)$ and $\mathbf{q}_3 = (0°, 120°)$; and case (c) is based on the optimal configurations $\mathbf{q}_1 = (0°, -57°)$, $\mathbf{q}_2 = (0°, 0°)$ and $\mathbf{q}_3 = (0°, 57°)$. As follows from these results, here the proposed approach allows us to increase the robot accuracy by 18% compared to D-optimal plan and by 56% compared to non-optimal plan of experiments.

## 6  Application example: calibration of Kuka KR270

Now let us present a more sophisticated example that deals with calibration experiments design for the industrial robot KUKA KR-270 (Fig. 4a). The geometrical model and parameters of the robot are presented in Fig. 4b [13]. For this case study, the parameters $d_0$, $d_5$, $\Delta q_6$ do not affect the robot accuracy. For this reason, they are eliminated from the list of parameters used in the experiment design.

Accordingly, the optimization problem (15) associated with the calibration experiment plans for $m \in \{3, 4, 12\}$ has been solved. While solving this problem, it was assumed that the end-effector position is estimated using FARO laser tracker (Fig 3c) [14], for which the measurement errors can be presented as unbiased random values with s.t.d. $\sigma = 0.03$ mm. For the computations, the workstation Dell Precision T7500 with two processors Intel® Xeon® X5690 (Six Core, 3.46GHz, 12MB Cache12) and 48 GB 1333MHz DDR3 ECC RDIMM has been used. Since the optimisation problem (15) is quite sensitive to the starting point, parallel computing with huge number of the initial points were used. To increase robustness of the proposed approach, the starting points were selected taking into all constraints. Besides, filtering of the points that correspond to the high values of $\rho_0$ has been applied.

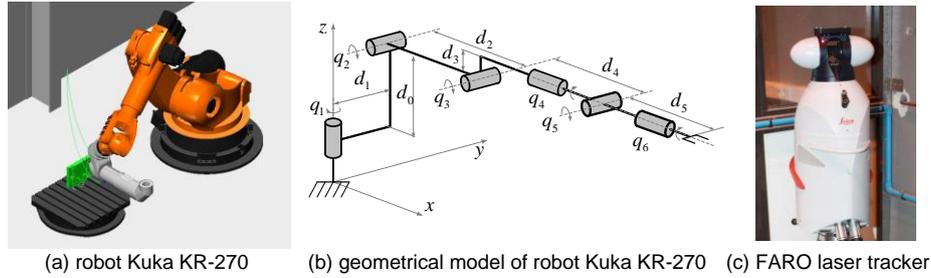

(a) robot Kuka KR-270    (b) geometrical model of robot Kuka KR-270    (c) FARO laser tracker

**Fig. 4.** Robot Kuka KR-270, its geometrical model and FARO laser tracker

The obtained results and comparison study with random plan are summarized in Table 2. Here, random plans have been generated 20 000 times using joints and workspace limits. Table 2 includes maximum, minimum and mean values of the performance measure $\rho_0$ for the generated sets of configurations. It has been shown that within the proposed plan of experiments, the calibration is much more efficient and high accuracy can be achieved using 3-4 measurement configurations only. Table 2 also includes some additional results obtained by multiplication of the measurement configurations, which show that it is not reasonable to solve optimization problem for 12 configurations (that produce 72 design variables). However, almost the same accuracy of the error compensation can be achieved by carrying out 12 measurements in 4 different configurations only (3 measurements in each configuration).

For comparison purposes, Fig 5 presents simulation results obtained for different types of calibration experiments. Here, each point corresponds to a single calibration experiment with random measurement errors. As follows from the obtained results, any optimal plan (obtained for the case of three, four or twelve calibration experiments) improves the accuracy of the compliance error compensation in the given test pose by about 75% comparing to the random plan. Also, it is shown that repeating experiments with optimal plans obtained for the lower number of experiments provides almost the same accuracy as the "full-dimensional" optimal plan. Thus, this idea of the reduction of the measurement pose number looks very attractive for the engineering practice.

**Table 2.** Accuracy of the error compensation $\rho_0, [m \times 10^{-6}]$ for different plans of experiments.

| Number of experiments | | 3 | 4 | 3×4 | 4×3 | 12 |
|---|---|---|---|---|---|---|
| Random plan | max | 47.1×10⁶ | 8078 | 23.6×10⁶ | 2693 | 144 |
| | min | 101 | 76.2 | 50.0 | 44.0 | 44.3 |
| | mean | 0.49×10⁶ | 375 | 0.24×10⁶ | 217 | 67.2 |
| Proposed plan | | 63.7 | 52.1 | 31.9 | 30.1 | 30.0 |

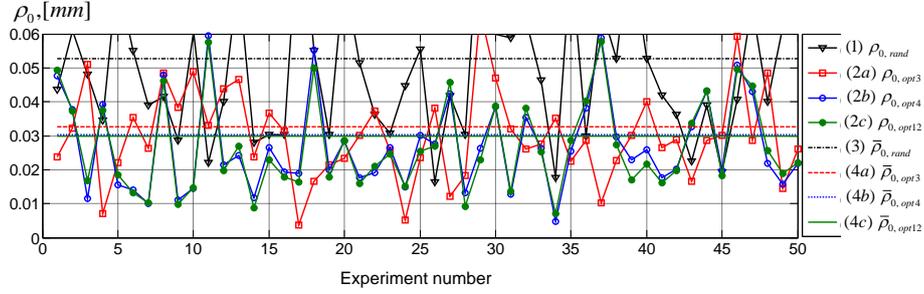

**Fig. 5.** The accuracy of errors compensation in the test configuration for different plans of calibration experiments for Kuka KR-270 robot for $\sigma = 0.03\,mm$ : (1) random plan $\rho_{0,\,rand}$ ; (2a) four experiments for optimal plan obtained for three calibration experiments $\rho_{0,\,opt3}$ , (2b) three experiments for optimal plan obtained for four calibration experiments $\rho_{0,\,opt4}$ , (2c) experiments for optimal plan obtained for twelve calibration experiments, $\rho_{0,\,opt12}$ ; (3) expectation for the plan (1) $\bar{\rho}_{0,\,rand} = 52.7 \cdot 10^{-3}\,mm$ ; (4a) expectation for the plan (2a) $\bar{\rho}_{0,\,opt3} = 32.7 \cdot 10^{-3}\,mm$ ; (4b) expectation for the plan (2b) $\bar{\rho}_{0,\,opt4} = 30.3 \cdot 10^{-3}\,mm$ ; (4c) $\bar{\rho}_{0,\,opt12} = 29.8 \cdot 10^{-3}\,mm$ ;

## 7 Conclusions

The paper presents a new approach for the design of calibration experiments for robotic manipulators that allows essentially reducing the identification errors due to proper selection of the manipulator configurations. In contrast to other works, the quality of the calibration experiment plan is estimated using a new performance measure that evaluates the efficiency of the error compensation in the given test-pose. This approach ensures the best position accuracy for the given test configuration.

The advantages of the developed technique are illustrated by two examples that deal with the calibration experiment design for 2 d.o.f. and 6 d.o.f. manipulators. The results show that the combination of the low-dimension optimal plans gives almost the same accuracy as the full-dimension plan. This heuristic technique allows user to reduce essentially the computational complexity required for the calibration experiment design. In a future work, an additional investigation will be performed for the experiment design for the set of the test poses (or for a long machining path).

**Acknowledgments.** The work presented in this paper was partially funded by the ANR, France (Project ANR-2010-SEGI-003-02-COROUSSO).

# References

1. Elatta, A.Y., Gen, L.P., Zhi, F.L., Daoyuan, Y., Fei, L.: An Overview of Robot Calibration. *Information Technology Journal*. 3, 74-78 (2004)
2. Veitchegger, W.K., Wu, C.H.: Robot accuracy analysis based on kinematics. *IEEE Journal of Robotics and Automation*. 2, 171-179 (1986)
3. Khalil W., Dombre E.: Modeling, identification and control of robots. Hermes Penton, London (2002)
4. Hollerbach J., Khalil W., Gautier M.: Chapter: Model identification, In: *Springer Handbook of robotics*. Springer. 321-344 (2008)
5. Atkinson A., DoneA.: Optimum Experiment Designs. Oxford University Press (1992)
6. Daney, D.: Optimal measurement configurations for Gough platform calibration. *IEEE International Conference on Robotics and Automation* (ICRA). 147-152 (2002)
7. Daney, D., Papegay, Y., Madeline, B.: Choosing measurement poses for robot calibration with the local convergence method and Tabu search. *The International Journal of Robotics Research*. 24, 501-518 (2005)
8. Klimchik, A., Wu, Y., Caro, S., Pashkevich, A.: Design of experiments for calibration of planar anthropomorphic manipulators. *IEEE/ASME International Conference on Advanced Intelligent Mechatronics* (AIM). 576-581 (2011)
9. Khalil, W., Gautier, M., Enguehard, Ch.: Identifiable parameters and optimum configurations for robots calibration. *Robotica*. 9, 63-70 (1991)
10. Sun, Y., Hollerbach, J.M.: Observability index selection for robot calibration. *IEEE International Conference on Robotics and Automation* (ICRA). 831-836 (2008)
11. Borm, J.H., Menq, C.H.: Determination of optimal measurement configurations for robot calibration based on observability measure. *J. of Robotic Systems*. 10, 51-63 (1991)
12. Imoto, J., Takeda, Y., Saito, H., Ichiryu, K.: Optimal kinematic calibration of robots based on maximum positioning-error estimation (Theory and application to a parallel-mechanism pipe bender). *Proc. of the 5th Int. Workshop on Computational Kinematics*. 133-140 (2009)
13. KUKA Industrial Robots, www.kuka-robotics.com/
14. FARO Laser Tracker, http://www.faro.com/lasertracker/